\documentclass{article}

\PassOptionsToPackage{numbers,sort&compress}{natbib}

     \usepackage[final]{tackling_climate_workshop_style}

\usepackage[utf8]{inputenc} % allow utf-8 input
\usepackage[T1]{fontenc}    % use 8-bit T1 fonts
\usepackage{hyperref}       % hyperlinks
\usepackage{url}            % simple URL typesetting
\usepackage{booktabs}       % professional-quality tables
\usepackage{amsfonts}       % blackboard math symbols
\usepackage{nicefrac}       % compact symbols for 1/2, etc.
\usepackage{microtype}      % microtypography
\usepackage{authblk}
\usepackage{amsmath} 
\usepackage{amsfonts}
\usepackage{graphicx}
\usepackage{xcolor}
\usepackage{enumitem}
\usepackage[belowskip=-10pt,aboveskip=0pt]{caption}
\usepackage{subcaption}

\usepackage{tabularx}
\usepackage{wrapfig}

\title{Learning to forecast vegetation greenness at fine resolution over Africa with ConvLSTMs}
\author[1]{\textbf{Claire Robin}}
\author[1]{\textbf{Christian Requena-Mesa}}
\author[1]{\textbf{Vitus Benson}} 
\author[1]{\textbf{Lazaro Alonso}}
\author[1]{\textbf{Jeran Poehls}}
\author[1,2]{\\ \textbf{Nuno Carvalhais}}
\author[1,2]{\textbf{Markus Reichstein}}
\affil[1]{Biogeochemical Integration, Max-Planck-Institute for Biogeochemistry, Jena, Germany}
\affil[2]{ELLIS Unit Jena, Michael-Stifel-Center Jena for Data-driven and Simulation Science, Jena, Germany}
\affil[ ]{Corresponding author: \textit
{\{crobin, crequ\}@bgc-jena.mpg.de}}

\begin{document}

\maketitle

\begin{abstract}
    Forecasting the state of vegetation in response to climate and weather events is a major challenge. Its implementation will prove crucial in predicting crop yield,  forest damage, or more generally the impact on ecosystems services relevant for socio-economic functioning, which if absent can lead to humanitarian disasters. Vegetation status depends on weather and environmental conditions that modulate complex ecological processes taking place at several timescales. Interactions between vegetation and different environmental drivers express responses at instantaneous but also time-lagged effects, often showing an emerging spatial context at landscape and regional scales. 
    We formulate the land surface forecasting task as a strongly guided video prediction task where the objective is to forecast the vegetation developing at very fine resolution using topography and weather variables to guide the prediction.
    We use a Convolutional LSTM (ConvLSTM) architecture to address this task and predict changes in the vegetation state in Africa using Sentinel-2 satellite NDVI, having  ERA5 weather reanalysis, SMAP satellite measurements, and topography (DEM of SRTMv4.1) as variables to guide the prediction. 
    Ours results highlight how ConvLSTM models can not only forecast the seasonal evolution of NDVI at high resolution, but also the differential impacts of weather anomalies over the baselines. The model is able to predict different vegetation types, even those with very high NDVI variability during target length, which is promising to support anticipatory actions in the context of drought-related disasters \footnote{https://github.com/earthnet2021/earthnet-models-pytorch.git}.

\end{abstract}

\section{Introduction}
Climate change is leading to an increase in extreme weather events, affecting both ecosystem and human livelihoods. Africa is one of the most vulnerable continents to climate change with droughts in the region expected to increase in severity according to the IPCC \citep{portner2022climate}. 
%During the past 20 years, droughts have directly caused the death of more than 20,000 people and affects the life of almost 270 billion. 
Given current projections of population growth and impacts from climate change, many more people will be affected by extreme drought events in the future \citep{yaghmaei2019disasters}. Any insight into predicting their occurrences can help prepare short-term solutions to alleviate potential impacts on local and regional communities \citep{mehrabi2019can}. 

The local scale response to extreme events is often not homogeneous \citep{kogan1990remote}. Abiotic (\textit{e.g.} soil type, topography, water bodies) and biotic (\textit{e.g.} vegetation type, plant rooting depth) factors and the interactions between them affect how vegetation responds to atmospheric extreme events \citep{pelletier2015forecasting, sturm2022satellite}.In addition, weather impacts with temporal lagged responses can have a significant influence on ecosystem responses to weather variability \cite{ogle2015quantifying, papagiannopoulou2017non, johnstone2016changing}. This so-called 'memory effect', arises from the complex processes involved in vegetation dynamics leading to non-linear interaction on several time scales \citep{de2015model, kraft2019identifying} (\textit{e.g.} the time-lagged effect of precipitation on vegetation due to the water soil recharge \citep{seddon2016sensitivity}, legacy effect of earlier drought \citep{bastos2021vulnerability, sturm2022satellite} or early starting season \citep{sippel2016ecosystem}). 

Predicting the evolution and impacts of vegetation at the very local level is therefore a challenge to support anticipatory action before a disaster 
In this paper, we aim to create a model capable of learning the relationship between vegetation states, local factors and weather conditions, at a very fine resolution in order to characterise and forecast the impact of weather extremes from an ecosystem perspective. We use the Normalized Difference Vegetation Index (NDVI) \citep{tucker1979red} as a proxy for vegetation health monitoring. Our work can be easily extended to predict other vegetation indices, depending on the ecological process of interest. 

\paragraph{Contributions} Our main contributions are summarized as follows:
\begin{itemize}[noitemsep,topsep=0pt]
    \setlength{\itemsep}{0pt}%
    \setlength{\parskip}{0pt}
    \renewcommand\labelitemi{\small$\bullet$}
    %A new dataset for landsurface forecasting on the African continent.
    \item A proof-of-concept of forecasting vegetation greenness in Africa at high spatial resolution.
    \item A spatio-temporal analysis of the prediction, both of the dataset and of individual samples.
\end{itemize}

\paragraph{Application context}
%We propose a model to forecast vegetation evolution at the very local level is to develop an early warning system. It is an effective way, encouraged by IPCC, for disaster risk reduction, social protection programs, and managing risks to health and food systems \citep{portner2022climate}, \citep{cutter2012managing}. 

Anticipatory action, such as Forecast-based Financing (FbF) \citep{Perez2014ForecastbasedFA}, implement short-time action (e.g. commercial animal destocking or early procurement of food) during the period of time between a warning and a disaster to reduce both the impact of the disasters and the financial cost of humanitarian aid  \citep{mechler2005cost}. One of the main obstacles to anticipatory action is the uncertainty of a disaster actually taking place, what is its magnitude and where and when exactly respond \citep{hillbruner2012early} \cite{maxwell20122011}. This context leads to protracted debates about response strategy and a reluctance to make decisions on the part of donors and policymakers during these precious time windows \citep{Perez2014ForecastbasedFA}. 

%Our model predict at local scale the impact of extreme weather from an ecosystem perspective, taking into account the specificity of each ecosystem. 
Since drought vulnerability is very context specific and location specific \citep{gonzalez2016learning}, a forecast vegetation evolution tool predicting at the very local scale from an ecosystem perspective can be impactful for the drought vulnerability assessments in term of location and magnitude to support anticipatory action. Additionally, our model is relevant for area not well covered by \textit{in situ} meteorological and vegetation monitoring instruments (which is notably the case in Africa \citep{easterling2013global}) since we use only on satellite data.

\paragraph{Related work}
At low spatial resolutions, machine learning models have been proposed for both vegetation modeling \citep{koehler2020forecasting,gobbi2019high,das2016deep,fan2021gnn,cui2020forecasting, kraft2019identifying, kang2016forecasting, lees2022deep} and crop forecasting \citep{kamilaris2018deep,schauberger2020systematic,khaki2020cnn}. \citet{requena2021earthnet2021} introduced Earth surface forecasting as modeling the future spectral reflectance of the Earth surface and provided the first dataset in this respect, EarthNet2021. Concurrent to our work, \citet{diaconu2022understanding} and \citet{kladny2022deep} have  built ConvLSTM variants for EarthNet2021. Both study the influence of weather on the predictions by providing artificial meteorological inputs.

 \begin{figure*}
    \centering
    \hspace*{-0.4cm} 
    \includegraphics[width=\textwidth]{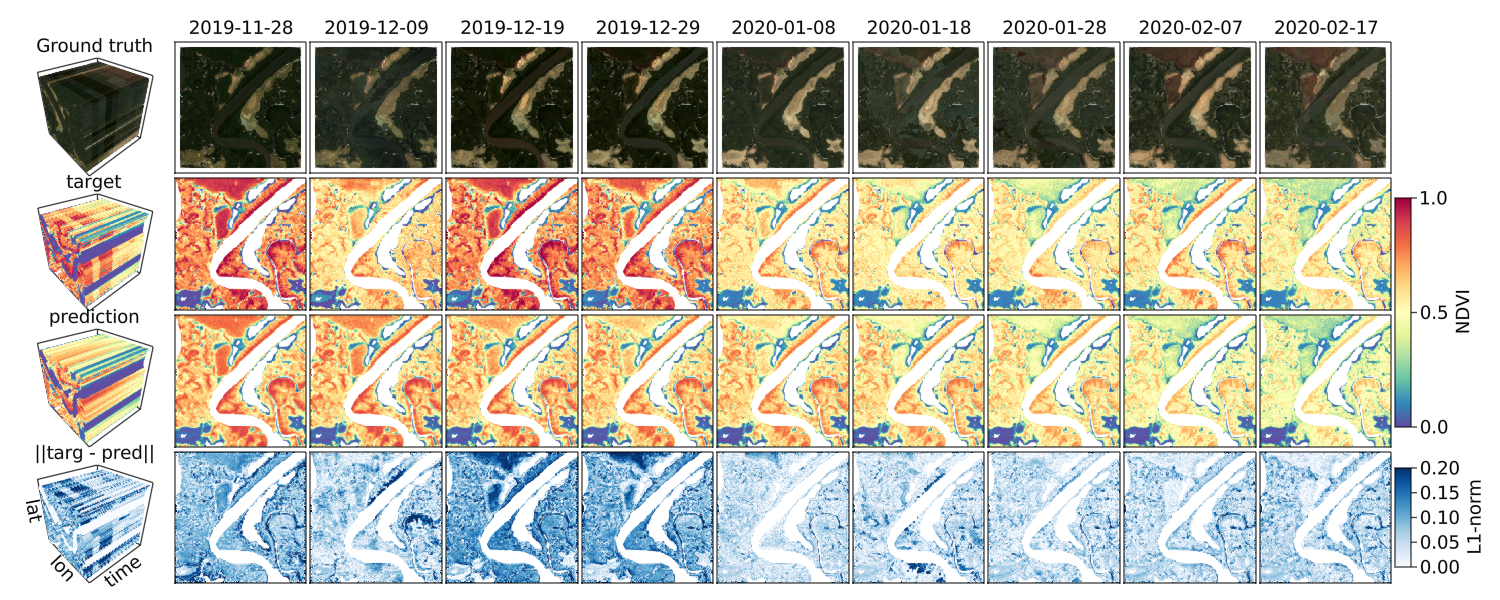}
    \caption{\textbf{Model prediction frame by frame}. The frames are temporally interpolated in a spatio-temporal cube \textbf{(Left side)}. \textbf{(Top row)} RGB satellite imagery. \textbf{(Second row)} NDVI target. \textbf{(Third row)} ConvLSTM predictions. \textbf{(Bottom row)} L1-norm of the difference between the target and the prediction. \textbf{Location:} 13°16'34.8"N 16°07'10.4"W, The Gambia.}
    \label{fig:EF_archi}
\end{figure*}

\section{Method}
This paper follows the works of \citet{requena2021earthnet2021}, who introduced EarthNet2021, focusing instead on Africa and a specific deep learning model: the Convolutional LSTM (ConvLSTM).

\paragraph{Dataset}
Similar to EarthNet2021 \citep{requena2021earthnet2021}, our dataset contains Sentinel 2 satellite imagery (bands red, blue, green and near-infrared), weather variables from ERA5 reanalysis (evaporation, surface pressure, surface net solar radiation, 2m temperature, total precipitation, potential evaporation) \citep{hersbach2020era5} and SMAP (latent and sensible heat flux, rootzone and surface soil moisture, surface pressure) \citep{entekhabi2010soil} and topography from SRTMv4.1 DEM \citep{reuter2007evaluation}. The data is collected at high spatial resolution for over $40000$ locations in Africa, each resulting sample we call a \textit{minicube}. For more information, see supplementary materials \ref{datadesc} and \ref{datasources}.

\paragraph{Task} We define an Earth surface forecasting task \citep{requena2021earthnet2021} as a strongly guided video prediction task. The objective is to forecast a length-k sequence of future NDVI satellite imagery ($\widehat{S_{n + 1}}, ..., \widehat{S_{n + k}}$) based on previous length-n sequence of satellite imagery ($S_{1}, ..., S_n$), topography $T$ and guiding environmental variables during both the context and prediction periods ($E_1, ..., E_n, ...,  E_{n + k}$). Formally, the task is to learn a function $f$ such that:

\begin{equation*}
    (\widehat{S_{n + 1}}, ...,\widehat{S_{n + k}}) = f(S_{1}, .., S_{n}, E_{1}, ...,E_n, ..., E_{n + k}, T) 
\end{equation*}

In this paper, deep learning models use a context period of one year and then forecast the next three months of NDVI at a $10$-daily timestepping. 
 
\paragraph{Model}
We use a ConvLSTM \citep{shi2015convolutional}, that is an LSTM \citep{hochreiter1997long} using convolution to operate in the spatial domain. We stack two ConvLSTM units (each designed as in \citet{patraucean2015spatio}) into an encoder and another two into a decoder (for forecasting). The encoder is used during the context period, it gets as inputs the concatenated satellite imagery, environmental variables and topography. The decoder uses only the environmental variables and the topography as inputs, but gets the hidden states from the encoder to propagate information from the context period. The model is visualized in supplementary material fig.~\ref{fig:AF}. The training procedure is detailed in supplementary material \ref{training}.

\textbf{Baselines and ablation} We compare our model against a \textit{constant baseline} (last valid pixel from context period) and a \textit{previous season baseline} (observations from previous year). Furthermore we ablate our model by removing the environmental variables (\textit{ConvLSTM w/o weather}), i.e. doing an ablation without weather to see how much the model learns from just the first order process underlying vegetation dynamics: the seasonal cycle.  
 
\textbf{Evaluation} We evaluate the models using two mean squared error ($MSE$) derived scores: the root mean squared error ($RMSE$) and the Nash-Sutcliffe model efficiency ($NSE$) \citep{nash1970river}. The latter rescales the $MSE$ with the variance of the observations $\sigma_0^2$, it is defined as:
\begin{equation}
    NSE = 1 - MSE/\sigma_0^2,
\end{equation}

Both the $MSE$ and the $NSE$ can be decomposed into parts:
\begin{align}
    MSE &= phase\_err + var\_err + bias\_sq \\
    NSE &= 2 \alpha r - \alpha^2 - \beta^2,
\end{align}
Here we introduced the bias $bias\_sq = (\mu_0 - \hat{\mu})^2$, the variance error $var\_err = (\sigma_0 - \hat{\sigma})^2$, the phase error $phase\_err = (1-r)*2* \sigma_0 * \hat{\sigma}$, the correlation $r$, the rescaled bias $\beta = (\hat{\mu} - \mu_0)/\sigma_0$ and a measure of relative variability $\alpha = \hat{\sigma}/\sigma_0$. $\mu$ represents means, $\sigma^2$ represents variances, $\cdot_{0}$ refers to observations and ${\hat{\cdot}}$ stands for predicted quantities. For the $NSE$ decomposition, the components' ideal values are $r = 1$, $\alpha = 1$ and $\beta = 0$ \citep{gupta2009decomposition}.

\section{Results}

\begin{table}
    \centering
    \begin{tabularx}{\textwidth}{Xccccc}
     \toprule
          Model & $RMSE$ $\downarrow$ & $NSE$ $\uparrow$ & $\alpha$ & $\beta$ & $r$  \\
          %ideal value & $0$ & $1$ & $1$ & $0$ & $1$\\
          %&\textit{mean} & \textit{median} & \textit{median} & \textit{median} & \textit{median}\\
        \midrule
        Constant baseline & 0.3365 & -1.3922 & 0.0 & 0.1559 & 0.0 \\
        Previous season baseline & 0.2937 & -1.0561 & 1.0169 & -0.0084 & 0.5504\\
        ConvLSTM without weather & 0.2331 & -0.3356 & 0.6512 & 0.1699 & 0.7348 \\
        ConvLSTM (ours) & \textbf{0.1882} & \textbf{0.0270} & 0.7570 & 0.0628 & 0.8024\\
    \bottomrule
    \end{tabularx}
    \vspace*{4pt}
    \caption{\textbf{Test set model performance.} All values are computed on minicubes with less than $75\%$ masked values, shown are the median values over all pixels. We additionally exclude the $5\%$ worst pixels as outliers because the metrics are unbounded.
    } %of{table}
    \label{tab:MSE_lstm}
\end{table}

\begin{figure}
    %\hspace*{-0.7cm}
    \includegraphics[width=\textwidth]{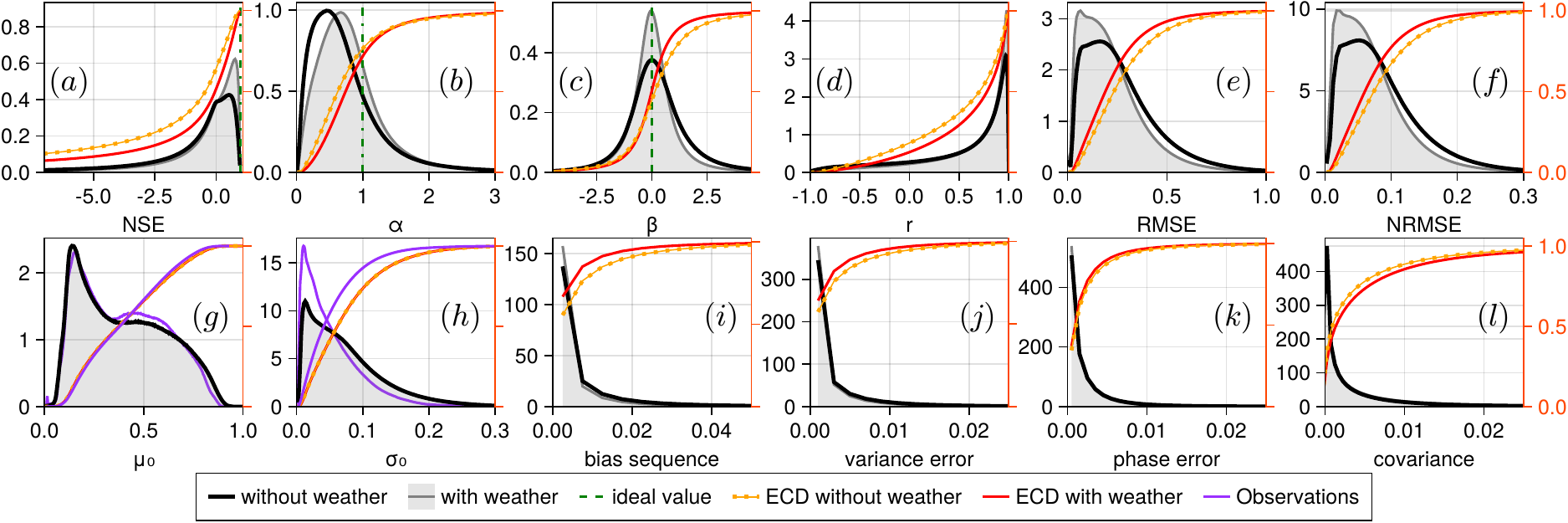}
    \caption{\textbf{Probability density plots of pixelwise test set performance.} ECD is the empirical cumulative distribution function. \textbf{(First row)} NSE decomposition. Performance distributions are closer to ideal values for $NSE$ $\alpha$, $\beta$ and $r$ with weather than without it. The same pattern is observed for $RMSE$, normalized $RMSE$ ($NRMSE$) and the decomposition of $MSE$ into bias sequence, variance error and phase error.}
    \label{fig:histo}
\end{figure}

%\begin{wrapfigure}{R}{0.75\textwidth}
%\end{wrapfigure}
For our evaluation, we removed minicubes with more than $75\%$ missing pixels. The evaluation is computed per pixel of the time series: we compute the metrics for each unmasked pixel of each minicube. Central estimates of our models' performance are reported in tab.~\ref{tab:MSE_lstm}. Our ConvLSTM model beats the baselines by a large margin in both $RMSE$ and $NSE$. The previous season baseline has the same NDVI distribution over the whole dataset as the target (since it is the same season one year ago), so we have $\sum \hat{\sigma} \simeq \sum \sigma_0$ and $\sum \hat{\mu} \simeq \sum \mu_0$. Negative errors compensate for positive ones, which explains a median of $\alpha$ and $\beta$ very close to their ideal value (but not for $r$). An ablation of it not using weather covariates performs worse than our model using them. This supports the intuition that weather should be driving vegetation dynamics. Supplementary figs.~\ref{fig:scatters} and~\ref{fig:pvs_season} visualize the performance against the baselines clustered by land cover type. Our model is stronger than the baselines across all classes.

%We further decipher the results by decomposition of residuals. 
Fig.~\ref{fig:histo} shows the decomposition for both $NSE$ and $RMSE$ on the test set. We find $50\%$ of the pixels in our model reach $NSE \ge 0$, while this is only $35\%$ in the ablation without weather. This difference mostly stems from a reduction in bias (fig.~\ref{fig:histo}c) and a reduction in relative variability (fig.~\ref{fig:histo}b).

While already good, our ConvLSTM does have limitations. Fig.~\ref{fig:EF_archi} visualizes a prediction on one example minicube. One visible limitation is due the dataset. On 2019-12-09, the gapfilled RGB satellite image has noise due to undetected clouds. Clouds lower the target NDVI and lead to an apparent overestimation of the model on that day. Additionally, since such artifacts are common in the training dataset, the model learns a biased estimate, slightly underestimating NDVI to account for occasional clouds. Such underestimation is visible on 2019-12-19 in fig.~\ref{fig:EF_archi}. We present the pixelwise decomposition of the $NSE$ for this pixel in the supplementary material, fig.~\ref{fig:nse}.

\section{Conclusion}
This work presents a ConvLSTM deep learning model to predict vegetation greenness in Africa at high spatial resolution from coarse-scale weather. Our model obtains a higher Nash-Sutcliffe model efficiency than two baselines. In an ablation we train a model variant without weather covariates with lower performance. Hence, our final model is able to extract information from meteorology. We trained our model on a diverse dataset, thus it is applicable to a wide variety of landcover classes and climate zones. Decomposing the NSE into parts, we find a vulnerability of our model to data noise: some clouds were not flagged during pre-processing, thereby biasing our model towards lower greenness values. Nevertheless, our model is a proof-of-concept of high resolution vegetation modeling in Africa. Future work should now focus on improving predictions by better understanding of spatial context, on the applicability of our model during extreme events and on building a bridge to the anticipatory action community.

\paragraph{Authors contribution}
\textbf{C.R}: Manuscript, experimental design, visualisation. \textbf{C.R-M}: Supervision, manuscript, dataset consititution. \textbf{V.B}: dataset constitution, pre-processing, pytorch framework, manuscript. \textbf{L.A}: Visualization, manuscript revision, model evaluation. \textbf{N.C}: Supervision, manuscript revision. \textbf{M.R}: Supervision, manuscript revision.

\begin{ack}
\textit{This project has received funding from the European Union’s Horizon 2020 research and innovation programme under grant agreement No 101004188.}
\end{ack}

\bibliography{main}

\section{Supplementary material}

\begin{figure*}[h!]
    \centering
    \includegraphics[scale=0.39]{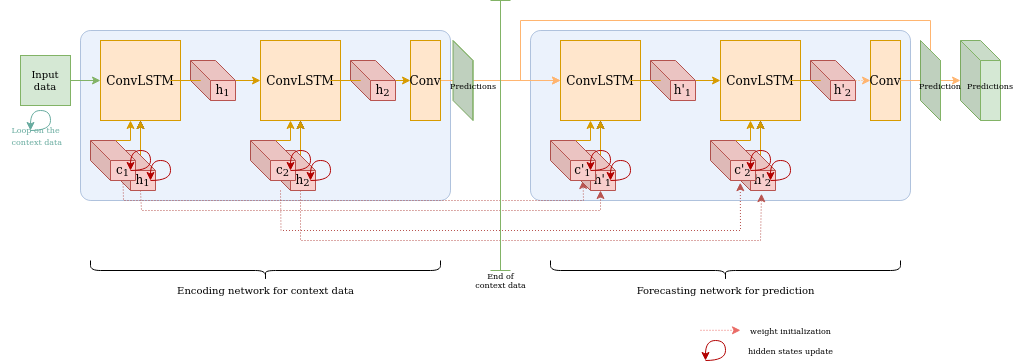}
    \caption{\small{\textbf{An Encoding-forecasting architecture}. The encoding network is a stack of two ConvLSTM, the forecasting network is a stack of two ConvLSTM. The hidden states of the first network are initialized at zero, the hidden states of the second network are initialized with the the hidden states' values of the first network. We use skip connections between two predictions, to predict only the difference. The first network is useful to to store as much information as possible during the context period while the second focuses on prediction..}}
    \label{fig:AF}
\end{figure*}

\begin{figure}[h!]
    \centering
    \includegraphics[scale=0.35]{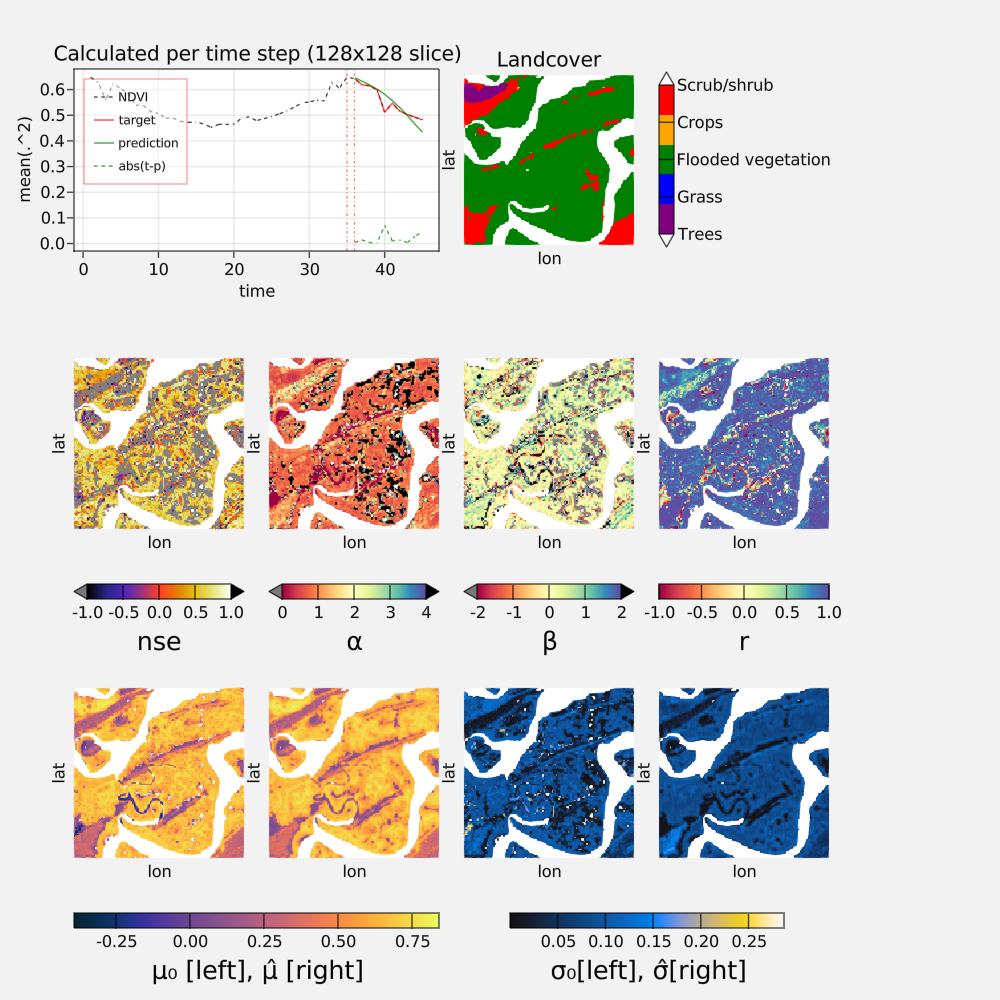}
     \caption{\textbf{Nash-Sutcliffe Efficiency decomposition} \textbf{(Top left)} Calculation of the average of each frame, to study the temporal variability. Before the red lines, it is the context, after, the prediction. \textbf{(Top right)} Landcover. \textbf{Bottom} NSE decomposition, the NSE is per pixel, computed on the time series.}
    \label{fig:nse}
\end{figure}

\begin{figure}[h!]
  \centering
\includegraphics[width=\textwidth]{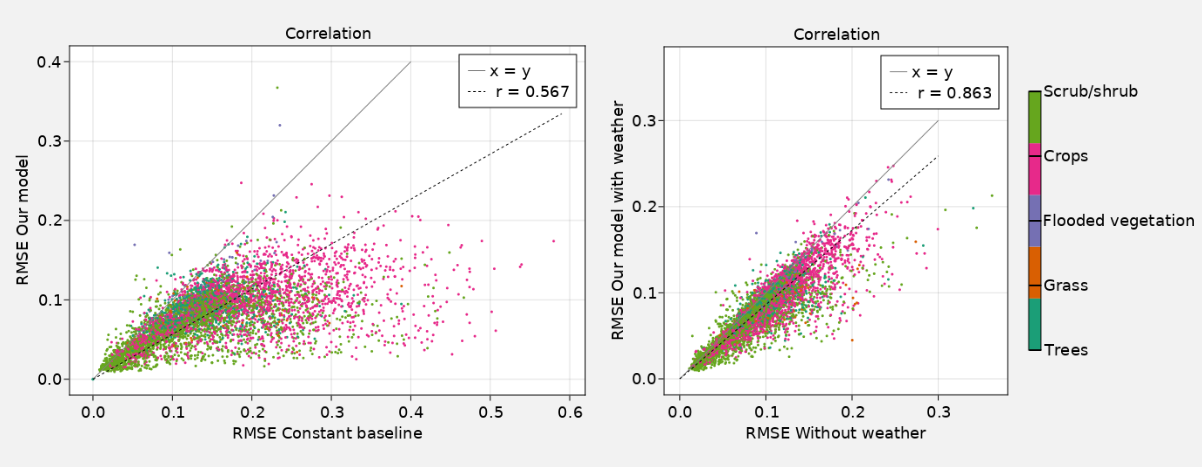}
\caption{\textbf{Correlation between the error of our model against the ablation study and the constant baseline}. The x-axis is the error of the ablation study, the y-axis is the error of our model. Each point below equation x = y is better predicted by our model, each point above is less well predicted. \textbf{(Right) Correlation between the error prediction of our model and the constant baseline}. Constant baseline predicts constantly the last NDVI values seen during the context period. The worse the baseline performance, the greater the variation in NDVI during the target period, and thus the more difficult the prediction task can be considered.\textbf{(Left) Correlation between the error prediction of our model and the ablation study without weather information}.}
\label{fig:scatters}
\end{figure}
\begin{figure}[h!]
    \centering
    \includegraphics[scale=0.45]{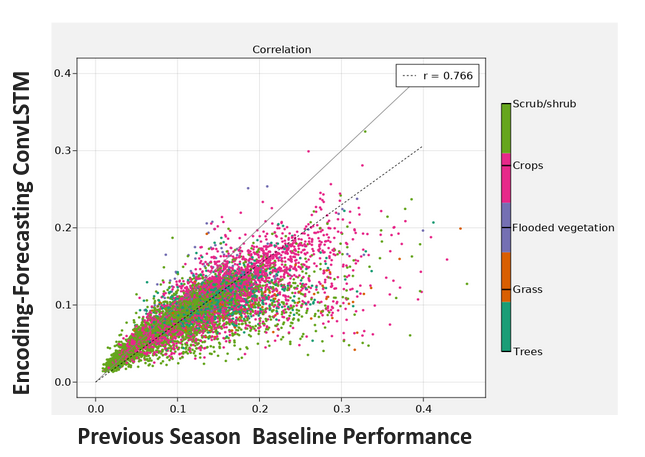}
    \caption{\textbf{Correlation between the error prediction of our model and the previous season baseline}. Previous baseline predicts the NDVI values of the previous season. The x-axis is the error of the baseline, the y-axis is the error of our model. The worse the baseline performance, the greater the evolution of the NDVI is different from the previous year, which may be due to different weather conditions.}
    \label{fig:pvs_season}
\end{figure}

\subsection{Dataset description} \label{datadesc}
Following the EarthNet2021 challenge \citep{requena2021earthnet2021}, we design a dataset for landsurface forecasting task on the African continent. 
The dataset contains over 40,000 samples, each sample is called a $minicubes$ given that they are three dimensional arrays. The temporal resolution is 10 days, with 45 time-step. There is 36 time-step per year and each minicube contains 5 seasons. The data come from 2016 to 2021, with a start in four different dates: first of March, June, September or December. Minicubes's length is 15 months. The resolution of the spatial data is around 30 meters per pixels, so an image of with a size of $128 \times 128$ pixels represent an area of $3.84 \times 3.84$ km. Minicubes in the same location never overlap temporally, but the same location can be in the test and the train set (for different years). Auto-correlation between different nearby locations is a limitation of the data set. We chose to randomly split the years in the training and test sets due to interannual climate variability, e.g. El Niño. Splitting the years with only a few years can potentially add a potential out-of-domain problem for training, we preferred to simplify this problem in this proof of concept. Therefore, the years are considered independent. Half of the samples are dominated by croplands, the other half is dominated by other land cover classes present in Africa (e.g. forest, mangroves, savanna, meadows, scrub, shrubs or bare land).

\subsection{Data sources} \label{datasources}
Each minicube contains 45 time-step of high resolution Sentinel-2 satellite imagery in the red, blue, green,  and near-infrared, so respectively the B02, B03, B04 and B08 reflectance bands \citep{drusch2012sentinel}. We additionally compute the Normalized Difference Vegetation Index (\textbf{NDVI}) \citep{tucker1979red}, a vegetation index (VI) for vegetation health monitoring based on the red and near-infrared bands. This is the target for the landsurface forecasting task.
Each minicube contains the \textbf{topography} of the location based on a digital elevation model (DEM) of the Shuttle Radar Topography Mission (SRTMv4.1) \citep{reuter2007evaluation}. The original spatial resolution is 90 meters, we use interpolation to get a 30 meters resolution for each of the $128 \times 128$ pixels.
For the weather, we use the reanalysis dataset ERA5 \citep{hersbach2020era5}. We use \textbf{observational weather} to reduce both the uncertainty and the computation cost of a seasonal weather forecasting, the only information available when used for real forecasts. We use the following weather variables: surface pressure, surface net solar radiation, 2m temperature, potential evaporation, and total precipitation. We additionally use the SMAP satellite measurements of the \textbf{land surface soil moisture} \citep{reichle2019version}, the soil moisture surface and soil moisture rootzone variables. Finally, each minicube also contains the ESRI2020 \textbf{landcover map} \citep{karra2021global}. We use the landcover to detect the non-vegetation pixels (water, build area, bare ground, snow or cloud) and mask them during training and evaluation, we also use it for results analysis, but not during the training. 

\subsection{Pre-processing}
 The data has been processed using the Level-2A algorithm \citep{muller2013sentinel} to the cloud detection following by interpolation on the cloudy data. Then we apply a custom timeseries jump filter based on \citep{chen2004simple} but adapted for the sentinel-2 data to reduce the noise due to the undetected clouds.
 However, the pre-processing is imperfect, due to the difficulty of correctly detecting clouds and their shadows. This issue leads to artifacts in the data. The undetected clouds distort the subsequent calculation of NDVI, additionally, when the interpolation is made from undetected cloud images, the result propagates the clouds in the following time steps.

\subsection{Training procedure} \label{training}
The loss function is the L2-norm. We apply during training and evaluation a mask to remove all the missing and non-vegetation pixels based on the landcover map. The target is the NDVI and the prediction is done timestep by timestep recurrently. When weather and environmental variables are in input of the ConvLSTM, their dimensions are upscaled to $128 \times 128$, moreover, future weather is on input even during the prediction period since it is a strongly guided task. The topography is attached to each timestep. We train our models for 150 epochs with a batch size of 32 and a learning rate of $10^{-6}$. Models were implemented using the deep learning framework PyTorch Lightning \cite{falcon2020pytorchlightning} which is built on top of PyTorch \cite{paszke2019pytorch}.

\subsection{Limitation of the evaluation}
The accuracy of the model for different seasons and for latitudes and longitudes depends strongly on the type of vegetation (agricultural or natural vegetation), without showing any particular pattern. There is a strong correlation between target variance and prediction accuracy (i.e., the sample with the most variance, e.g., during the growing season, is more difficult to predict than those without variance), see Fig. \ref{fig:scatters} and Fig. \ref{fig:pvs_season}. In addition, samples with many clouds often have poor prediction. However, the limitation of the dataset prevents a finer prediction: many clouds have not been detected, and therefore it is difficult to attest the influence of clouds on the accuracy of the model. Moreover, we are working with a whole continent, containing several types of vegetation, climate and ecosystems, so the drivers are radically different \citep{seddon2016sensitivity}, as  as well as the time scales. Without any information on meteorological and vegetation anomalies, we cannot propose a more accurate evaluation at this point.

\end{document}